\documentclass[sigconf]{acmart}

\usepackage{amsmath}
\usepackage{url}
\usepackage{graphicx}
\usepackage{subcaption}
\usepackage{multirow}

\usepackage{algorithm}
\usepackage{algpseudocode}

\usepackage{amsmath}
\usepackage{xcolor}
\usepackage{booktabs}
\usepackage{comment}

\usepackage{textcomp}
\usepackage{siunitx}
\usepackage{enumitem}
\usepackage{todonotes}

\usepackage{mathtools}

\DeclareMathOperator*{\argminC}{\arg\min}   

\newcommand\ourmethod{AdaFair~}

\def\BibTeX{{\rm B\kern-.05em{\sc i\kern-.025em b}\kern-.08emT\kern-.1667em\lower.7ex\hbox{E}\kern-.125emX}}
    
\begin{document}
\fancyhead{}
%
\title{AdaFair: Cumulative Fairness Adaptive Boosting}

\author{Vasileios Iosifidis}
\email{iosifidis@L3S.de}
\affiliation{%
  \institution{L3S Research Center, Leibniz University of Hannover}
  \city{Hannover}
  \country{Germany}
}
\author{Eirini Ntoutsi}
\email{ntoutsis@L3S.de}
\affiliation{%
  \institution{L3S Research Center, Leibniz University of Hannover}
  \city{Hannover}
  \country{Germany}
}


%

\begin{abstract}
The widespread use of ML-based decision making in domains with high societal impact such as recidivism, job hiring and loan credit has raised a lot of concerns regarding potential discrimination. In particular, in certain cases it has been observed that ML algorithms can provide different decisions based on sensitive attributes such as gender or race and therefore can lead to discrimination. Although, several fairness-aware ML approaches have been proposed, their focus has been largely on preserving the \emph{overall} classification accuracy while improving fairness in predictions for both protected and non-protected groups (defined based on the sensitive attribute(s)). The overall accuracy however is not a good indicator of performance in case of class imbalance, as it is biased towards the majority class. As we will see in our experiments, many of the fairness-related datasets suffer from class imbalance and therefore, tackling fairness requires also tackling the imbalance problem.

To this end, we propose \ourmethod, a fairness-aware classifier based on AdaBoost that further updates the weights of the instances in each boosting round taking into account a cumulative notion of fairness based upon all current ensemble members, while explicitly tackling class-imbalance by optimizing the number of ensemble members for balanced classification error. 
Our experiments show that our approach can achieve parity in true positive and true negative rates for both protected and non-protected groups, while it significantly outperforms existing fairness-aware methods up to 25\% in terms of balanced error.
\end{abstract}

%



%

%
\maketitle

\section{Introduction}
\label{sec:intro}
AI-based decision making systems are employed nowadays in an ever growing number of application areas such as finance industry, autonomous driving, health-care, risk assessment, etc. The impressive performance of such systems - in several tasks, has reached or even outperformed human performance - leads to anticipation about automating human tasks by AI in the foreseeable future~\cite{grace2018will}. Such a demand for automation is also driven by the ever growing amount and complexity of data. 

However, in parallel to  the growing demand for automation using AI, 
an increased number of concerns regarding accountability, fairness and transparency of such systems, esp. for domains of high societal impact, has been raised over the recent years~\cite{united2014big}. 
There is already a plethora of observed incidents regarding discrimination caused by AI-based decision making systems~\cite{AmazonPrime,datta2015automated,airbnb,sweeney2013discrimination,larson2016we}.
In Amazon's Prime case~\cite{AmazonPrime}, for example, in which AI algorithm's task was to decide which areas of a city are eligible to advanced services, areas mostly inhabited by black people were ignored (\textit{racial-bias}), even though the algorithm did not consider race as a feature. In another case, it was discovered that Google's AdFisher tool displayed significantly more advertisements of highly paid jobs to men than women~\cite{datta2015automated} (\textit{gender-bias}). 
Such incidents point out the urge to consider fairness in AI algorithms in order to exploit the amazing potential of the technology for societal advance.

A growing body of works has been proposed over the last year to address the problem of fairness and algorithmic discrimination. Such methods can be broadly categorized into pre-, in- and post-processing approaches based on whether they ``correct'' for fairness at the input training data, at the algorithm itself or at the output model.
To the best of our knowledge, the vast majority of these methods, e.g.,~\cite{krasanakis2018adaptive,zafar2017fairness,calders2009building,calmon2017optimized,kamiran2012data,hardt2016equality,fish2016confidence,kamiran2009classifying,kamiran2010discrimination}, focus on optimizing for fairness while maintaining an overall high classification performance. Considering overall classification error is problematic in case of class-imbalance as the performance of the model on the minority class is typically ignored and therefore, such methods end up with low TPR and high TNR rates. Such approaches might achieve fairness, as typically fairness is evaluated in terms of performance parity between the protected and non-protected groups, however the predictive performance of the model on the minority group is insufficient (low TPRs). 
Recent state-of-the-art approaches do not consider this problem, as we show in Section~\ref{sec:evaluation}; these methods achieve fairness by lowering TPRs for both groups.

Class-imbalance is an inherited problem of fairness - as we will see in our experiments, many of the datasets exhibit high class imbalance (c.f., Table~\ref{tbl:datasets}) and therefore, tackling fairness requires also tackling imbalance\cite{galar2012review}.
Our proposed approach, \ourmethod, overcomes this issue and achieves fairness while preserving high TPRs and high TNRs for both groups.
\ourmethod is based on AdaBoost and extends its instance weighting strategy for each round based on the thus far observed ensemble fairness. This way, in each round, the weak learner focuses on both hard classification examples (as in traditional boosting) and on the discriminated group per class. 
The discriminated group, per class, is identified dynamically at each boosting round and its effect on the instance weighting is evaluated based on a cumulative notion of fairness that considers the fairness behavior of the thus far built ensemble model in the particular round.  
We further prioritize the instances for training, by taking into account not only the error of the weak learner but also its confidence in the predictions per instance so that harder instances that lie further away from the boundary, are further boosted. 
Finally, at the end of the training phase, we select the best sequence of weak learners which achieves high performance and fairness. 

Our contributions are summarized as follows: i) we propose AdaFair, a fairness-aware boosting method that achieves parity between the two groups (thus achieving fairness) while maintaining high TPRs and TNRs (thus tackling class imbalance). ii) We define the notion of cumulative fairness for the ensemble and a dynamic group weighting schema for the per round discriminated group. iii) \ourmethod outperforms current state-of-the-art methods in terms of performance in a value range from 8\% to 25\% (depending on the severity of class-imbalance). iv) We show the superiority of our cumulative notion of fairness vs a  non-cumulative alternative. v) We show that including confidence in the weighting, results in more confident predictions for AdaFair.

The rest of the paper is organized as follow: in Section 2, we review the related work. In Section 3, we define the problem and describe basic concepts. Our \ourmethod is presented in Section 4, whereas the experimental evaluation is described in Section 5. Conclusions and outlook are summarized in Section 6.

\section{Related Work}
\label{sec:related}
\noindent \textbf{Fairness notions.} 
A growing body of fairness notions have been proposed over the recent years. Up to now, more than twenty fairness notions exist for fairness in classification~\cite{romei2014multidisciplinary,verma2018fairness}, however there is no specific arrangement on which fairness notion is universally suitable. One of the earliest measures of discrimination, the so-called \textit{statistical} or \textit{demographic parity}~\cite{kamiran2009classifying,kamiran2012data}, measures the percentage difference between non-protected and protected group which are assigned to a target class i.e., grant a loan application, etc. However, this definition only requires a balanced representation of both groups to a target class without controlling if the selected instances are qualified or not~\cite{dwork2012fairness}. On the other hand, a recent measure, called equal opportunity~\cite{hardt2016equality,verma2018fairness}, eliviates this pitfall by measuring percentage difference between true positive percentages of two groups. Although equal opportunity has potential, it does not capture the full picture therefore \textit{equalized odds}~\cite{hardt2016equality} (also called \textit{disparate mistreatment}~\cite{zafar2017fairness}) has been proposed. Equalized odds extends equal opportunity by considering the difference of true classified instances between protected and non-protected group in all classes. Although a plethora of fairness notions exist, in this work we employ equalized odds since it seems the most promising notion and also has been adopted by recent state-of-the-art methods~\cite{zafar2017fairness,hardt2016equality,krasanakis2018adaptive,pleiss2017fairness}.

\noindent \textbf{Pre-processing approaches.} 
One of the most common causes of machine learning discrimination is arising from discrimination which resides in historical data. Pre-processing methods aim to deal with this issue by performing data transformations, perturbations or augmentation to eliminate underlying discrimination. These methods are typically model-agnostic and therefore, any classifier is applicable after the pre-processing phase. \textit{Massaging}~\cite{kamiran2009classifying}, \textit{re-weighting}~\cite{calders2009building}, \textit{uniform- or preferential-sampling}~\cite{kamiran2012data} and \textit{data augmentation}~\cite{iosifidisdealing} are pre-processing methods which change data distributions directly; they aim to restore balance among protected and non-protected group within data by changing instance labels, assigning different weights, under- or over-sampling instances, and generating pseudo-instances, respectively. \textit{Massaging} and \textit{re-weighting} have also been extended for stream classification~\cite{iosifidis2019fairness}. Another interesting approach by~\cite{calmon2017optimized}, performs data transformations to eliminate existing dependence of instances' labels. 

\noindent \textbf{In-processing approaches.} 
Approaches which operate during learner's training phase aim to mitigate discrimination as part of the objective function by employing set of constraints or using regularization. In~\cite{kamiran2010discrimination,zhang2019faht}, for example, they embed the statistical parity fairness notion into the splitting criterion of a decision tree. A regularization approach~\cite{kamishima2012fairness} scales down the correlation (mutual information) of the sensitive features and the class label in order to avoid outcomes based on these features. \cite{dwork2012fairness} introduced the notion of ``individual fairness-constraint" where similar instances should be treated similarly and then minimize a loss function subject to this notion. 
In \cite{zafar2017fairness}, they insert a set of convex-concave constraints, which aim to minimize equalized odds, into logistic regression classifiers. Finally,~\cite{krasanakis2018adaptive} assume that data does not contain robust and fair labels and proceed to estimate the true labels by re-adjusting iteratively the instance weights while minimizing equalized odds.

\noindent \textbf{Post-processing approaches.} 
Post-processing approaches can be divided into two sub categories: the ones that change the decision boundary of a model (white-box approaches) and the ones that directly change the prediction labels (black-box approaches). In the first category, for example~\cite{fish2016confidence} shift the decision boundary of AdaBoost to minimize discrimination while~\cite{DBLP:conf/sdm/PedreschiRT09} alter the confidence of CPAR classification rules and in~\cite{calders2010three} authors change Na\"ive Bayes probabilities w.r.t. fairness. 
Approaches in the latter category consider only the outcome of a classifier. For example,~\cite{hardt2016equality} set thresholds to predictions in order to achieve same error rates for protected and non-protected group. An extension of this work~\cite{pleiss2017fairness} analyze how to obtain calibrated classifiers under same error rates among groups. Finally, in~\cite{DBLP:conf/icml/AgarwalBD0W18} they build a fair classifier out of the predictions of another black-box classifier.

Since our approach can be considered as in-processing approach, we select methods from this category as competitors. We do not consider \cite{kamiran2010discrimination,kamishima2012fairness,dwork2012fairness} as competitors, while they are built upon fairness notions which seem incomplete or outdated and alternating them to optimize for another fairness notion may degrade their performance. Therefore, we select~\cite{zafar2017fairness,krasanakis2018adaptive} as our competitors while equalized odds is our common fairness measure. In contrast to our approach, our competitors only consider error rate in their loss function which fails to tackle the problem of class-imbalance.

\section{Basic concepts and definitions}
\label{sec:preliminaries}
We assume a dataset $D$ consisting of $n$ i.i.d. samples drawn from a joint distribution $P(F,S,y)$: 
$S$ denotes  %
\emph{sensitive attributes} such as gender and race, $F$ denotes other \emph{non-sensitive attributes} and $y$ is the class label.
For simplicity, we consider that the classification problem is binary, that is, $y \in \{+, -\}$ and that there exist a single sensitive attribute $S$, also binary. That is, $S \in \{s, \bar{s}\}$ with $s$ and $\bar{s}$ denoting the \emph{protected} and \emph{non-protected group}, respectively.
We use the notation $s_+$ ($s_{-}$), $\bar{s}_+$ ($\bar{s}_{-}$) to denote the protected and non-protected group for the positive (negative, respectively) class.
The goal of classification is to find a mapping function $f:(F,S) \rightarrow y$ to predict the class labels of future unseen instance. 

\textbf{Fairness definition:} As already discussed in Section~\ref{sec:related}, we adopt \emph{\textbf{Eq}ualized \textbf{Odds}} (shortly $Eq.Odds$)~\cite{hardt2016equality} as our fairness measure.
Eq.Odds, measures the difference in prediction errors between the protected and non-protected group. In particular, let $\delta FPR$ ($\delta FNR$) be the difference in false positive rates (false negative rates, respectively) between the protected and non-protected group, defined as follows ($\hat{y}$ denotes the predicted label):
\begin{equation}
\label{eq:FPR_FNR}
\delta FPR = P(y\neq \hat{y} | \bar{s}_-) - P(y\neq \hat{y} | s_-) \\
\delta FNR = P(y\neq \hat{y} | \bar{s}_+) - P(y\neq \hat{y} | s_+) 
\end{equation}

The goal is to minimize both differences, the so-called  $Eq.Odds$:
\begin{equation}
Eq.Odds = |\delta FPR |+ |\delta FNR|
\label{eq:EqOdds}
\end{equation}
The value range for each of $|\delta FPR|$, $|\delta FNR|$ is [0,1], where 0 stands for no discrimination and 1 stands for maximum discrimination. Eq.Odds values lie in [0-2] range.

The goal of fairness-aware classification is finding a mapping function $f(\cdot)$ that minimizes Eq.Odds discrimination while maintaining good predictive performance.
Typically, the predictive performance in the context of fairness is evaluated in terms of the \emph{error rate}, e.g.,~\cite{krasanakis2018adaptive,zafar2017fairness,calders2009building,calmon2017optimized,kamiran2012data,hardt2016equality,fish2016confidence,kamiran2009classifying,kamiran2010discrimination}, defined as:
\begin{equation}\label{eq:standard_error}
ER=\frac{FN+FP}{TP+TN+FN+FP}
\end{equation}

However, optimizing for error rate is problematic in cases of class imbalance. A possible outcome in such a case is that the classifier will misclassify most (in the extreme case all) of the minority instances while correctly classifying the majority - the error rate $ER$ will still be low despite the poor performance in the minority class.  
W.r.t. fairness such a classifier might still be fair, i.e., $Eq.Odds \approx 0$, as the difference between the FPRs, FNRs for each group will be low (c.f., Equations~\ref{eq:FPR_FNR},\ref{eq:EqOdds}).
However, the non-discriminative behavior of the classifier would be achieved by just reducing drastically the correct predictions for the minority class, with the extreme case of misclassifying everything from the minority.
Our goal in this work is to achieve $Eq.Odds$ while maintaining low FPRs (equivalently, high TNRs) and low FNRs (equivalently, high TPRs) for both groups.

To this end, we propose (c.f., Section~\ref{sec:AdaFair_tuning}) to replace the error rate (which is not a good performance indicator in case of imbalance, as we also show in Section~\ref{sec:evaluation}) with the \emph{balanced error rate} defined as follows~\cite{brodersen2010balanced}:
\begin{equation}\label{eq:balanced_error}
BER = 1 - \frac{1}{2}\cdot(\frac{TP}{TP + FN} + \frac{TN}{TN + FP})
= 1 - \frac{1}{2}\cdot(TPR + TNR)
\end{equation}

\textbf{AdaBoost:} 
Boosting is an ensemble technique that combines weak learners to create a strong learner. AdaBoost~\cite{schapire1999brief} calls a weak learner iteratively by adjusting the instance weights in each iteration (boosting round) based on misclassified instances.
We believe boosting is a promising technique for fairness-aware classification as it divides the learning problem into multiple sub-problems and then combines their solutions (sub-models) into an overall (global) model.
Intuitively, it is easier to tackle the fairness problem in the simpler sub-models rather than in a global complex model.
In order to apply AdaBoost for fairness one has to carefully change the underlying data distribution between consecutive rounds so that both predictive performance aspects and fairness-related aspects are considered (Section~\ref{sec:method}).


\section{Cumulative Fairness Adaptive Boosting}
\label{sec:method}
In this work, we tailor AdaBoost for fairness by adjusting its re-weighting process. In particular: i) we directly consider the fairness behavior of the model in the weighting process by introducing the notion of cumulative fairness that assesses the $Eq.Odds$ related behavior of the model up to the current boosting round (c.f., Equation~\ref{eq:cumulFair}). 
Moreover, differently from vanilla AdaBoost, ii) we employ confidence scores in the re-weighting process to allow for differentiation in instance weighting based on how confident is the model regarding their class.
The fairness-aware adjustments in the re-weighting process are described in Section~\ref{sec:AdaFair_reweighting}.  
Finally, we optimize the number of weak learners in the final ensemble by taking into account the balanced error rate and thus directly considering class imbalance in the best model selection (Section~\ref{sec:AdaFair_tuning}).

\subsection{Cumulative boosting fairness and fairness costs}
\label{sec:cumulativeFairness}
Let $j:1-T$ be the current boosting round, where $T$ is a user defined parameter indicating the number of boosting rounds.
Let $H_{1:j}(x) = \sum_{i=1}^j a_ih_i(x)$ be the ensemble model up to round $j$.

The \emph{cumulative boosting fairness} of the model $H_{1:j}(x)$ at round $j$ is defined in terms of $|\delta FPR|$, $|\delta FNR|$  for both protected and non-protected groups, as follows:

\begin{equation}\footnotesize
\label{eq:cumulFair}
\delta FNR^{1:j} = \frac{\sum_{i=1}^{|\bar{s}_+|} 1 \cdot\mathbb{I}[\sum_{k=1}^j a_kh_k(x_i^{\bar{s}_+}) \neq y_i]}{|\bar{s}_+|} - \frac{\sum_{i=1}^{|s_+|} 1 \cdot\mathbb{I}[\sum_{k=1}^j a_kh_k(x_i^{s_+}) \neq y_i]}{|s_+|}\\$$$$
\delta FPR^{1:j} = \frac{\sum_{i=1}^{|\bar{s}_-|} 1 \cdot\mathbb{I}[\sum_{k=1}^j a_kh_k(x_i^{\bar{s}_-}) \neq y_i]}{|\bar{s}_-|} - \frac{\sum_{i=1}^{|s_-|} 1 \cdot\mathbb{I}[\sum_{k=1}^j a_kh_k(x_i^{s_-}) \neq y_i]}{|s_-|}
\end{equation}\normalsize

where function $\mathbb{I}(\cdot)$ returns 1 iff the expression within is true, otherwise 0. In other words, the cumulative fairness at round $j$ evaluates current ensemble's parity among protected and non-protected groups for both positive and negative class. 

If there is no parity, we change the weights of the instances so that discriminated groups are  boosted extra in the next round $j+1$. Note that vanilla AdaBoost already boosts misclassified instances for the next round. Our weighting therefore aims at achieving parity across the groups and the classes. To avoid confusion we use the term \emph{costs} for our fairness-related extra weights.  
The fairness related costs are dynamically estimated in each round. 
In particular, the fairness related cost $u_i$ for an instance $x_i$ in the boosting round $j$ is computed as follows:   

\begin{equation}\footnotesize
\label{eq:fairnessCosts}
u_i = 
 \begin{cases}
 |\delta FNR^{1:j}|, & if~\mathbb{I}((y_i \neq h_j(x_i))\land |\delta FNR^{1:j}| >\epsilon), x_i \in s_+, \delta FNR^{1:j}> 0\\
 |\delta FNR^{1:j}|, & if~\mathbb{I}((y_i \neq h_j(x_i))\land |\delta FNR^{1:j}| >\epsilon), x_i \in \bar{s}_+,\delta FNR^{1:j}<0 \\
 |\delta FPR^{1:j}|, & if~\mathbb{I}((y_i \neq h_j(x_i))\land |\delta FPR^{1:j}| >\epsilon), x_i \in s_-, \delta FPR^{1:j}>0\\
 |\delta FPR^{1:j}|, & if~\mathbb{I}((y_i \neq h_j(x_i))\land |\delta FPR^{1:j}| >\epsilon), x_i \in \bar{s}_-,\delta FPR^{1:j}<0 \\ 0, & otherwise\\
 \end{cases}
\end{equation}\normalsize
where $u_i \in [0,1]$ and parameter $\epsilon \in \mathbf{R}$ reflects the tolerance to fairness and is typically set to zero or to a very small value.

At each round, instances which belong to groups that are treated unfairly, receive fairness-related cost. E.g., if in round $j$ group $s_+$ is discriminated, which means $\delta FNR^{1:j} >0$ and $\delta FNR^{1:j} > \epsilon$, then misclassified instances in this group will receive fairness related costs for the next round. The signs (+/-) of $\delta FNR^{1:j}$ and $\delta FNR^{1:j}$ denote which group will receive the costs in each class (only one group per class may receive fairness related costs), while $\epsilon$ is a condition for the necessity of fairness-related costs in the upcoming round $j+1$. Note that costs $u$ are global, i.e., all instances of a group and class combination will receive the same cost in a round $j$. 

\emph{Confidence scores:} AdaBoost weighting $a_j$ relies on the classification error of the weak learner $h_j$ built in round $j$. This is a global weighting, so all instances are weighted with the same $a_j$. Moreover, this weighting does not take into account how reliable the decision of the classifier was. To this end, we propose to also use the confidence of the predictions $\hat{h}_j(x)$ so that misclassified instances for which the learner is confident in its predictions receive more weight comparing to other instances.  
This differentiation in the weighting allows the classifier to focus on harder cases.

\subsection{The AdaFair Algorithm}
\label{sec:AdaFair_reweighting}

Algorithm~\ref{alg:method} shows AdaFair's training phase.
In the beginning, instance weights $w$ and costs $u$ are initialized (line 1). Next, a weak learner is trained upon a given weight distribution (line 2a) while the error rate, $\alpha_j$, $\delta FNR^{1:j}$, $\delta FPR^{1:j}$ and $u_i$ are computed for the current round (lines 2b, 2c, 2d, 2e and 2f, respectively).
After $u$ is computed, instance weights are estimated and normalized by a factor $Z$ (line 2g). In contrast to original AdaBoost, we employ the confidence score (line 2g, $\hat{h}$) to contribute to the re-weighting process. Based on the confidence score, we assign higher weights to misclassified instances which are harder to learn compared to instances which are near the decision boundary. 
The algorithm converges in 3 cases: a) $error = 0.5$, b) $Eq.Odds \leq \epsilon$ and $balanced~error \neq 0.5$ or c) maximum number of rounds is reached. 

\begin{algorithm}[t!]\small 
 \begin{flushleft}
 \hspace*{\algorithmicindent} \textbf{Input:} $D = (x_i,y_i)_1^N, T, \epsilon$ \\
 \hspace*{\algorithmicindent} \textbf{Output:} Ensemble $H$
 \end{flushleft}
 \caption{Training phase}
 \begin{enumerate}
 \item
 Initialize $w_i = 1/N$ and $u_i=0$, for $i = 1, 2, \dots, N$

 \item
 For $j = 1$ to T:
 \begin{enumerate}
 \item Train a classifier $h_j$ to the training data using weights $w_i$.

 \item
 Compute the error rate $\text{err}_j = \frac{\sum_{i=1}^N w_i I(y_i \neq h_j(x_i))}{\sum_{i=1}^N w_i}$

 \item Compute the weight  $\alpha_j = \frac{1}{2}\cdot \ln( \frac{1 - \text{err}_j }{ \text{err}_j})$
 \item Compute fairness-related $\delta FNR^{1:j}$ 

 \item Compute fairness-related  $\delta FPR^{1:j}$

 \item Compute fairness-related costs $u_i$
 
 \item Update the distribution as \\$w_i \leftarrow \frac{1}{Z_j} w_i \cdot e^{\alpha_j \cdot \hat{h}_j(x) \cdot \mathbb{I}(y_i \neq h_j(x_i))}\cdot (1 + u_i)$ \\// $Z_j$ is  normalization factor; $\hat{h}_j$ is the confidence score
 \end{enumerate}

 \item Output $H(x) = \sum_{j = 1}^T \alpha_i h_j(x)$
 \end{enumerate}
 \label{alg:method}
\end{algorithm}\normalsize 

\subsection{Optimizing the number of weak learners $\theta$}
\label{sec:AdaFair_tuning}
AdaBoost requires as input the number of training rounds $T$.  We propose to select the optimal number of weak learners $1 \cdots \theta, \theta \leq T$ that minimizes the error of the model. We propose to optimize for the balanced error rate (Equation~\ref{eq:balanced_error}) instead of the error rate $ER$ (Equation~\ref{eq:standard_error}) in order to tackle the imbalance problem and thus select a model that depicts good performance in both classes, i.e., high TPRs, high TNRs. 
In case of balanced data, BER corresponds to ER. To allow for different combinations of ER and BER in the $\theta$ computation, we consider both ER and BER in the objective function as follows: 
\begin{equation}\label{eq:argmin}
\argminC_\theta ~ (c\cdot BER_\theta + (1-c)\cdot ER_\theta + Eq.Odds_\theta)
\end{equation}
The parameter $c$ controls the impact of BER and ER in the computation. A detailed evaluation of its impact in the performance of \ourmethod is presented in Section~\ref{sec:exp_parameter_c}. As our analysis show, AdaFair achieves fairness (i.e., $Eq.Odds \approx 0$) for all different values of $c$, i.e., for all different combinations of ER and BER. However with $c=1$, that is when optimizing for BER, our method achieves the highest TPRs and only slight decreases in TNRs for both groups. Therefore, we recommend to use our method optimized for BER, that is, with $c=1$.
The result of this optimization step is a final ensemble model with $Eq.Odds$ fairness:
$H(x) = \sum_{i=1}^\theta a_ih_i(x)$.

\section{Evaluation}
\label{sec:evaluation}
The first goal of our experiments is to evaluate the predictive performance and fairness behavior of \ourmethod vs other related approaches (Section~\ref{sec:exp_comparison}).
Regarding predictive behavior, we report on both accuracy and balanced accuracy (Equation~\ref{eq:balanced_error}), whereas for fairness we report on \textit{Eq.Odds} (c.f., Section~\ref{sec:preliminaries}). Since \emph{Eq.Odds} (Equation~\ref{eq:EqOdds}) is an aggregated measure, we also report on TPR and TNR values for both protected and non-protected groups to shed more light on methods' performance. 
The second goal of our experiments is to understand the behavior of \ourmethod. To this end, we investigate the effect of cumulative vs non-cumulative fairness (Section~\ref{sec:single_vs_accum}) and the impact of adopting balanced error rate vs error rate (Section~\ref{sec:exp_parameter_c}). Moreover, we compare the cumulative distribution of margins to demonstrate the impact of confidence scores in the instance weight estimation (Section~\ref{sec:margins}).
Details on the datasets, baselines, parameter selection and evaluation are provided in Section~\ref{sec:expsetup}.

\subsection{Experimental setup}
\label{sec:expsetup}

\subsubsection{Datasets}
\label{sec:data}
We evaluate our approach on four real-world datasets whose characteristics are summarized in Table~\ref{tbl:datasets}. As we can see, they vary w.r.t. cardinality, dimensionality and class imbalance and therefore provide an interesting benchmark for evaluation.

\begin{table}[tp!]\scalebox{.8}{
\begin{tabular}{lcccc}
\hline
 & Adult Census  & Bank & Compass & KDD Census \\ \hline
\#Instances & 45,175 & 40,004 & 5,278 & 299,285 \\
\#Attributes & 14 & 16 & 9 & 41 \\
Sen.Attr. & Gender & Marit. Status & Gender &  Gender \\
Class ratio ($+$:$-$) & 1:3.03 & 1:7.57 & 1:1.12 & 1:15.11 \\
Positive class & \textit{\textgreater{}50K}  & \textit{subscription} & \textit{recidivism} & \textit{\textgreater{}50K}  \\ \hline
\end{tabular}}
\caption{An overview of the datasets.}
\label{tbl:datasets}
\end{table}

\noindent \textbf{Adult census income}~\cite{dua2017} dataset contains demographic data from the U.S. and the task is to predict whether the annual income of a person will exceed 50K dollars. 
The sensitive attribute is $S=Gender$ with $s=female$ being the protected group; the positive class is people receiving more than 50K. 
We remove duplicate instances and instances containing missing values. The positive to negative class ratio is ~1:3 (exact ratio 24\%:76\%).

\noindent \textbf{Bank} dataset~\cite{dua2017} is related to direct marketing campaigns of a Portuguese banking institution. The task is to determine if a person subscribes to the product (bank term deposit). As positive class we consider people who subscribed to a term deposit. We consider as $S=maritial~status$ with $s=married$ being the protected group. The dataset suffers from severe class imbalance, with a positive to negative ratio of ~1:8 (exact ratio 11\%:89\%).

\noindent \textbf{Compass} dataset~\cite{larson2016we} contains information on prisoners in Broward County such as the number of juvenile felonies. The task is to determine if a person will be re-arrested within two years \textit{(recidivism)}. We consider \textit{recidivism} as the positive class and $S=Gender$ with $s=female$ as the protected group. For this dataset, we followed the pre-processing steps of~\cite{zafar2017fairness}. The dataset is almost balanced, the exact positive to negative ratio is 46\%:54\%.

\noindent \textbf{KDD census income}~\cite{dua2017} has the same prediction task as the \textbf{adult census} dataset. However in KDD census ``the class labels were drawn from the total person income field rather than the adjusted gross income''~\cite{dua2017}.
We consider $S=Gender$ with $s=female$ as the protected group, and as positive class people receiving more than 50K annually. This is the most skewed dataset in our benchmark with a positive to negative ratio of ~1:15 (exact ratio 6\%:94\%).

\subsubsection{Baselines}
\label{sec:baselines}
We evaluate our approach against recently proposed state-of-the-art in-processing approaches that also aim to minimize  $Eq.Odds$, namely the methods of Krasanakis et al. and Zafar et al. 
In addition, we compare against two fairness-agnostic boosting versions: 
vanilla AdaBoost~\cite{schapire1999brief} and SMOTEBoost~\cite{chawla2003smoteboost} (an AdaBoost approach that tackles class-imbalance).
Finally, to study the behavior of our approach we also compare \ourmethod against different variations described hereafter:

\begin{enumerate}[wide, labelwidth=!, labelindent=0pt]

\item 
\textbf{Zafar et al.\cite{zafar2017fairness}:} The authors formulate the fairness problem as a set of convex-concave constraints to minimize $Eq.Odds$, for which they optimize a logistic regression model. 
\item 
\textbf{Krasanakis et al.~\cite{krasanakis2018adaptive}:} 
The authors assume the existence of latent fair classes and propose an iterative training approach towards those classes which alters in-training the instance weights. Fairness is assessed via $Eq.Odds$.  
\item 
\textbf{AdaBoost~\cite{schapire1999brief}:} 
This is the vanilla AdaBoost that does not consider fairness nor confidence scores. 
\item 
\textbf{SMOTEBoost~\cite{chawla2003smoteboost}:} This is an extension of AdaBoost for imbalanced data.
At each boosting round, new synthetic instances of the minority class are generated via SMOTE~\cite{chawla2002smote}, to compensate for the imbalance.
SMOTEBoost does not consider fairness nor confidence scores. The goal of this baseline is to see whether by only tackling imbalance the fairness problem can be addressed.  

\item 
\textbf{AdaFair:} 
Our proposed approach that combines cumulative fairness, balanced error rate and confidence scores.
\item 
\textbf{AdaFair NoCumul:}
Similar to \ourmethod, however the $Eq.Odds$ is computed per round rather than over all previous rounds, therefore impacting the instance weighting.
The goal of this baseline is to help clarifying the impact of cumulative vs non-cumulative fairness (Section~\ref{sec:single_vs_accum}).
\item 
\textbf{AdaFair NoConf:} 
Similar to \ourmethod, but is does not employ confidence scores for weight estimation. We employ this baseline to depict the impact of confidence scores in the cumulative distribution of margins of the training instances (Section~\ref{sec:margins}).
\end{enumerate}


\subsubsection{Parameter selection and evaluation}
We follow the same evaluation setup as in~\cite{zafar2017fairness,krasanakis2018adaptive} by splitting each dataset randomly into train $(50\%)$ and test set $(50\%)$ and report on the average of 10 random splits.
We set $\epsilon=0$ as a threshold for $Eq.Odds$, which means zero tolerance to discrimination. 
Moreover, we set the number of boosting rounds for \ourmethod to $T=200$
(same for the other ensemble approaches, c.f., Section~\ref{sec:baselines}). For Krasanakis et al. and Zafar et al. methods, we employ their default parameters. For SMOTEBoost, we set $N$ (the number of synthetic instances generated per round) to 2, 100, 100 and 500 for datasets compass, adult census, bank and kdd census, respectively. Furthermore, for experiments in Section~\ref{sec:exp_comparison}, \ref{sec:single_vs_accum} and \ref{sec:margins}, we set parameter $c=1$ (c.f., Equation~\ref{eq:argmin}), that is the proposed \ourmethod optimized for balanced error rate; the effect of $c$ is studied in Section~\ref{sec:exp_parameter_c}.

Our method\footnote{\url{https://iosifidisvasileios.github.io/AdaFair}} has been instantiated with Decision Stumps as weak learners. 

\subsection{Predictive and fairness performance}
\label{sec:exp_comparison}
\subsubsection{Adult census income}
\begin{figure}[tp!]
 \centering
 \includegraphics[width=.75\columnwidth]{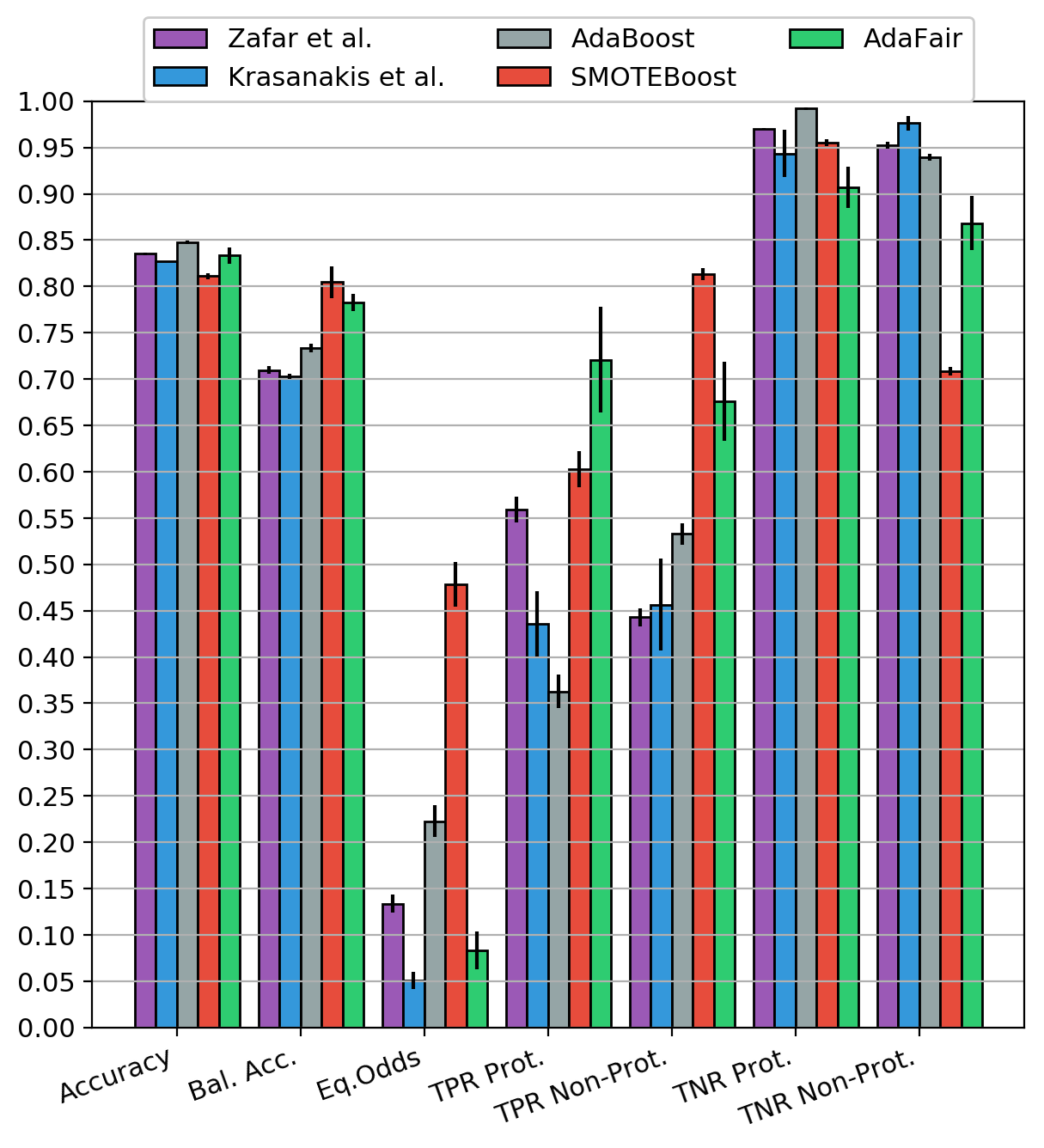}
 \caption{Adult census: Predictive and fairness performance - higher values are better; for Eq.Odds, lower values are better.}
 \label{fig:adult_performance}
\end{figure}

In Figure~\ref{fig:adult_performance}, we show the performance of the different approaches on Adult census dataset. 
Regarding predictive performance, we report on accuracy and balanced accuracy ($Bal.Acc.$ for short) whereas regarding fairness, we report on $Eq.Odds$ and $TPR$, $TNR$ values for both protected and non-protected groups ($Prot.$ and $Non-Prot.$, respectively for short).

The best balanced accuracy is achieved by SMOTEBoost followed by \ourmethod ($2\%\downarrow$); both methods target class imbalance, the latter however also considers fairness. 
AdaBoost, Krasanakis et al. and Zafar et al. that do not consider class imbalance have a 5\%$\downarrow$, 8\%$\downarrow$ and 8\%$\downarrow$, respectively drop in their balanced accuracy comparing to \ourmethod.

Regarding $Eq.Odds$, as expected AdaBoost and SMOTEBoost perform worse as they do not consider fairness. The best overall $Eq.Odds$ score is achieved by the method of Krasanakis et al., followed by our \ourmethod ($3\%\uparrow$, recall that lower values are better). A closer look, however, at the actual TPRs and TNRs values per group shows that our method achieves the highest TPRs values for both protected and non-protected groups compared to the other two fairness aware approaches. In particular, for the protected (non-protected) group our TPR is 29\%$\uparrow$ (22\%$\uparrow$, respectively) higher that of the second best method of Krasanakis et al. 
So, it seems that the methods of Krasanakis et al. and Zafar et al. produce low TPRs and high TNRs, i.e., these methods ``reject'' more instances of the positive class in order to minimize $Eq.Odds$ (this explains their high TNRs, low TPRs values). On the contrary, our \ourmethod achieves good performance for both classes (high TPRs, high TNRs) while maintaining a good $Eq.Odds$ (i.e., low difference in TPRs, TNRs for both protected and non-protected group).

\subsubsection{Bank}
\label{sec:bank_results}
\begin{figure}[tp!]
 \centering
 \includegraphics[width=.75\columnwidth]{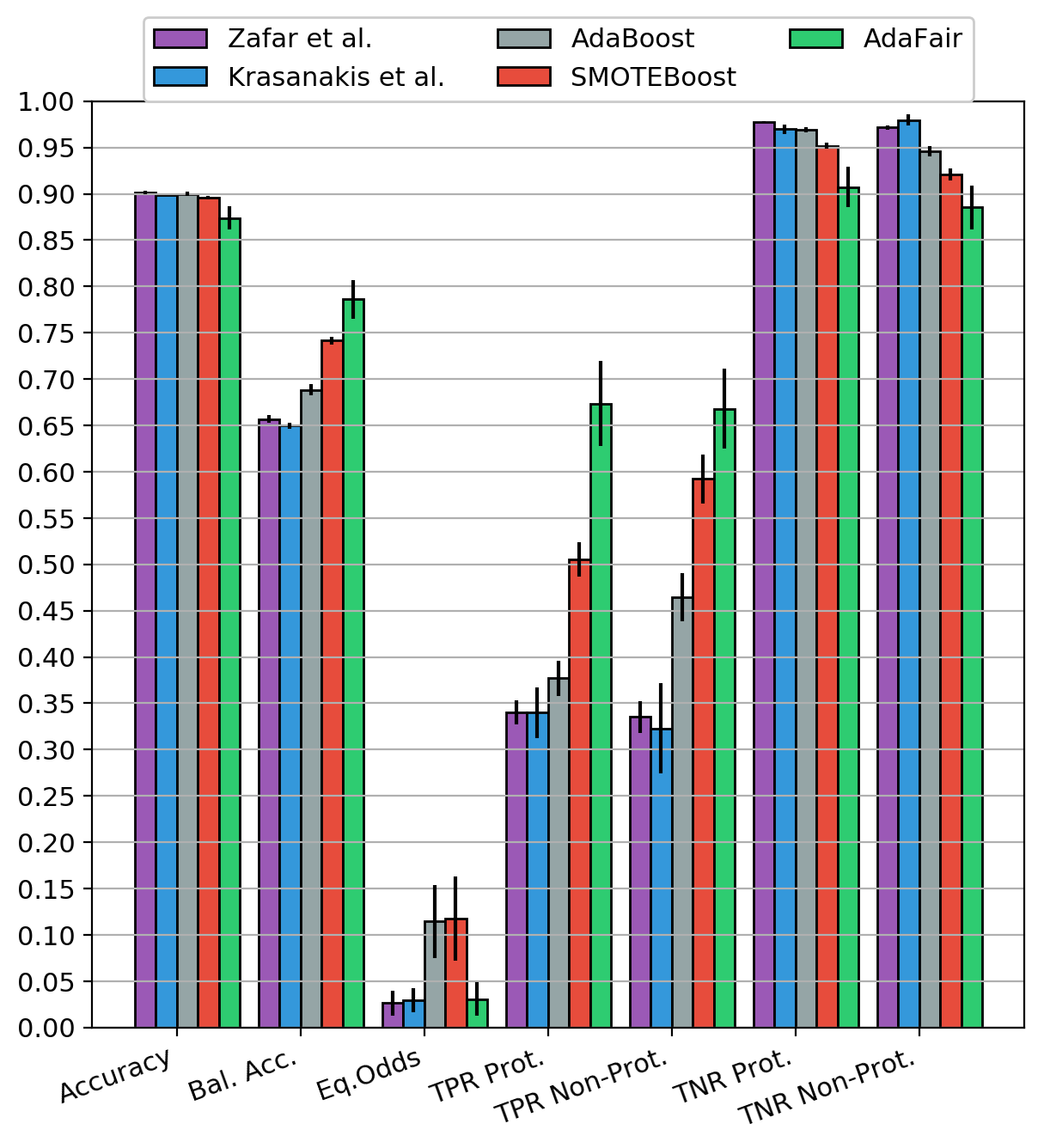}
 \caption{Bank: Predictive and fairness performance - higher values are better; for Eq.Odds, lower values are better.}
 \label{fig:bank_performance}
\end{figure}

The results are shown in Figure~\ref{fig:bank_performance}. All approaches, except for AdaBoost and SMOTEBoost, achieve low $Eq.Odds$. Interestingly, AdaFair achieves better balanced accuracy than SMOTEBoost, while it outperforms the other approaches.

A closer look at $Eq.Odds$ and namely at TPRs and TNRs shows significant differences between the approaches.
Namely, w.r.t. TPRs our method outperforms the second best (Zafar et al.) by almost 33\%$\uparrow$ for each group. Interestingly, AdaBoost and SMOTEBoost maintain higher TPRs than the methods of Krasanakis et al. and Zafar et al., even though they do not consider fairness. The methods of Krasanakis et al. and Zafar et al. have very similar behavior and it seems that both of them focus on the majority class (therefore high TNRs, low TPRs). 
Regarding TNRs, our method has a small drop of 6\%$\downarrow$ and 7\%$\downarrow$ drop for the protected and non-protected group compared to the second best approach of Zafar et al.; this is expected as we optimize for balanced error rather than overall error.

\subsubsection{Compass}
\label{sec:compass_results}
\begin{figure}[tp!]
 \centering
 \includegraphics[width=.75\columnwidth]{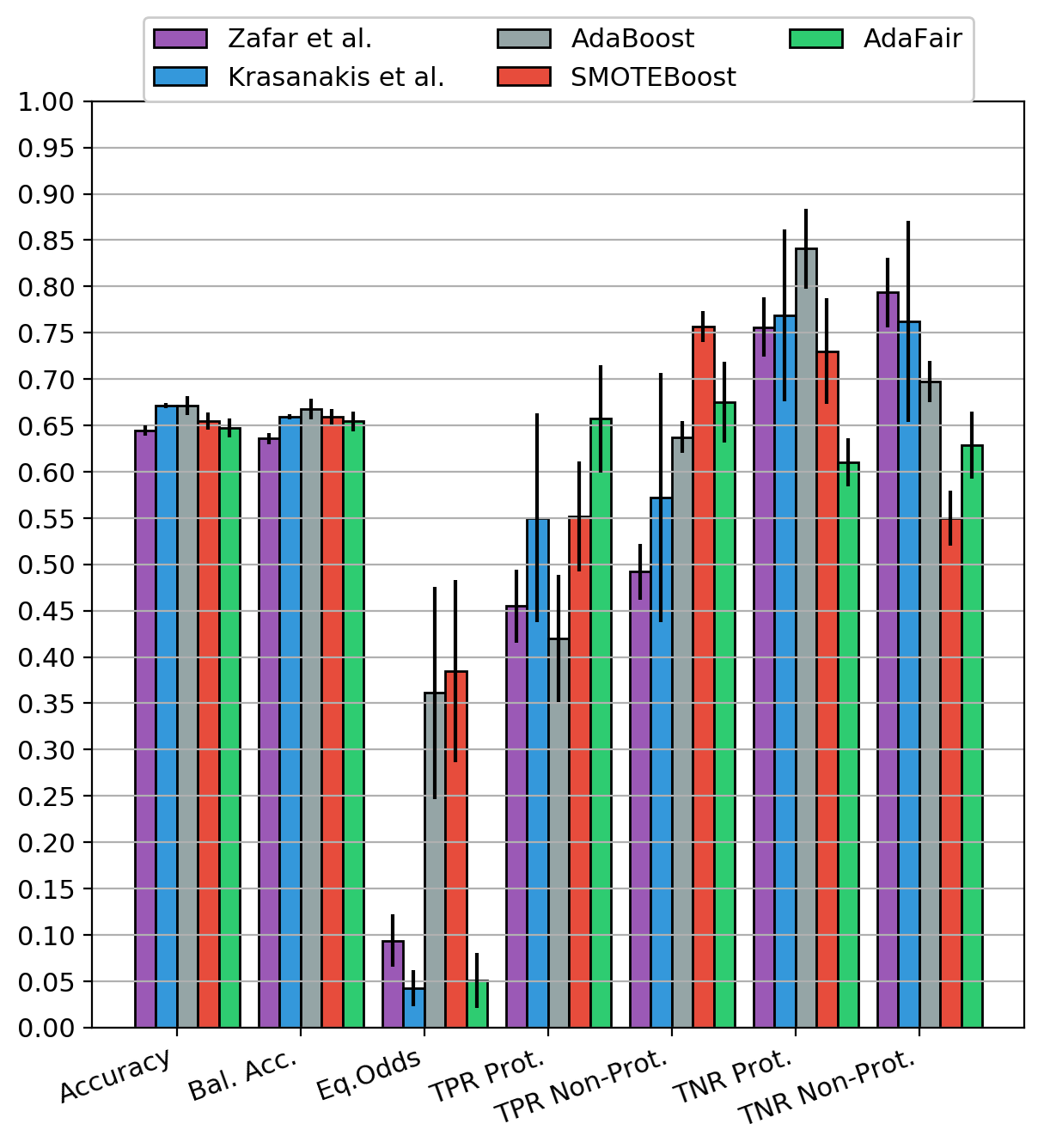}
 \caption{Compass: Predictive and fairness performance - higher values are better; for Eq.Odds, lower values are better.}
 \label{fig:compass_performance}
\end{figure}

The results are shown in Figure~\ref{fig:compass_performance}.
Regarding balanced accuracy, AdaBoost performs best and Zafar et al worse. However, the differences between the approaches are not that high. The  similar performance of the different approaches is to be expected as the dataset is balanced (c.f., Table~\ref{tbl:datasets}) and therefore, imbalance treatment has no strong effect.

Regarding fairness, the method of Krasanakis et al. achieves slightly better performance in terms of $Eq.Odds$ (0.6\%$\downarrow$) and balanced accuracy (0.2\%$\uparrow$) compared to the second best  \ourmethod. 
Zafar et al. has the worst $Eq.Odds$ (almost twice the value of Krasanakis et al.), recall that its $Bal.Acc.$ was the worse among the approaches.

By examining the TPRs and TNRs of both protected and non-protected groups, we observe that the performance of Krasanakis et al. is not stable (highest standard deviation among the methods). Our \ourmethod has better TPR values for both groups. Our TNRs are the lowest among the approaches ($~61\%-65\%$) as we optimize for balanced error and the negative class represents 54\% of the population. 

\subsubsection{KDD census income}
\label{sec:kdd_results}
\begin{figure}[tp!]
 \centering
 \includegraphics[width=.75\columnwidth]{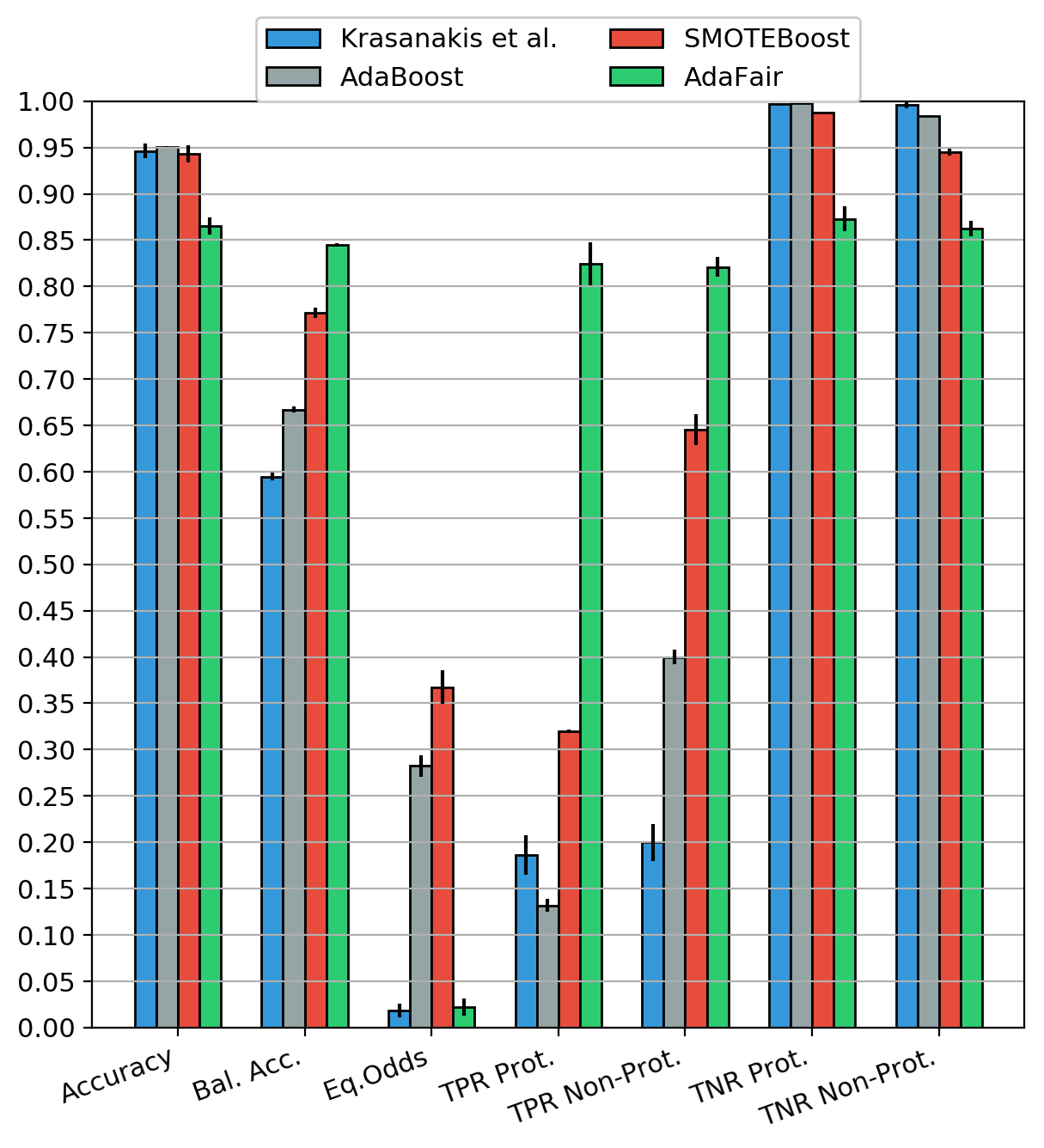}
 \caption{KDD census: Predictive and fairness performance - higher values are better; for Eq.Odds, lower values are better.}
 \label{fig:kdd_perormance}
\end{figure}
The dataset suffers from extremely high class imbalance, with a ratio of 1:15 (c.f., Table~\ref{tbl:datasets}). Zafar et al. approach could not be applied to this dataset due to its inability to estimate the optimal parameters, therefore we omit it.

In terms of balanced accuracy \ourmethod performs 25\%$\uparrow$ than Krasanakis et al., 18\%$\uparrow$ than AdaBoost and 8\%$\uparrow$ than SMOTEBoost. AdaBoost and Krasanakis et al. classify almost perfectly the negative class (i.e., TNRs close to 100\%), which comprises 94\% of the population. SMOTEBoost has 2\%$\downarrow$ to 4\%$\downarrow$ drop in TNRs of protected and non-protected group respectively, compared to AdaBoost.
In terms of TPR, \ourmethod achieves the highest TPR scores for both groups (above 80\%) while the method of  Krasanakis et al. results in values below 20\%. 
Both fairness-aware approaches, \ourmethod and Krasanakis et al., minimize discrimination to 2\% while AdaBoost and SMOTEBoost result in 28\% and 36\% $Eq.Odds$, respectively. 

\noindent\textbf{Conclusion:} To conclude, our \ourmethod is able to achieve high balanced accuracy and low discrimination by maintaining high TPRs and only slightly worse TNRs for both groups. On the contrary, the other fairness-aware approaches, namely Zafar et al. and Krasanakis et al, eliminate discrimination by reducing TPRs, that is by rejecting more instances of the positive class to achieve parity among the protected and non-protected groups. Moreover, Zafar et al. is unable to handle multi-dimensional datasets while it can not estimate the optimal parameters, therefore failing to be applied to multi-dimensional datasets. Furthermore, we perform paired t-test between \ourmethod and each baseline for all datasets. The results show highly significant difference ($p <0.001$) for all datasets between \ourmethod and each baseline (it is also visible from the reported TPRs and TNRs of each dataset).

\subsection{Cumulative vs non-cumulative fairness}
\label{sec:single_vs_accum}
The notion of cumulative fairness (c.f. Equation~\ref{eq:cumulFair}), is crucial for \ourmethod' ability to mitigate discrimination.
To investigate its impact we compare our \ourmethod (with cumulative fairness of models $1:j$, where $j$ is the current boosting round) with a version that 
considers only the fairness of the individual weak learner at round $j$ (refereed to as \ourmethod NoCumul).
 
\begin{figure*}[htp!]
 \centering
 \begin{subfigure}[t]{0.24\textwidth}
 \centering
 \includegraphics[width=1.04\columnwidth]{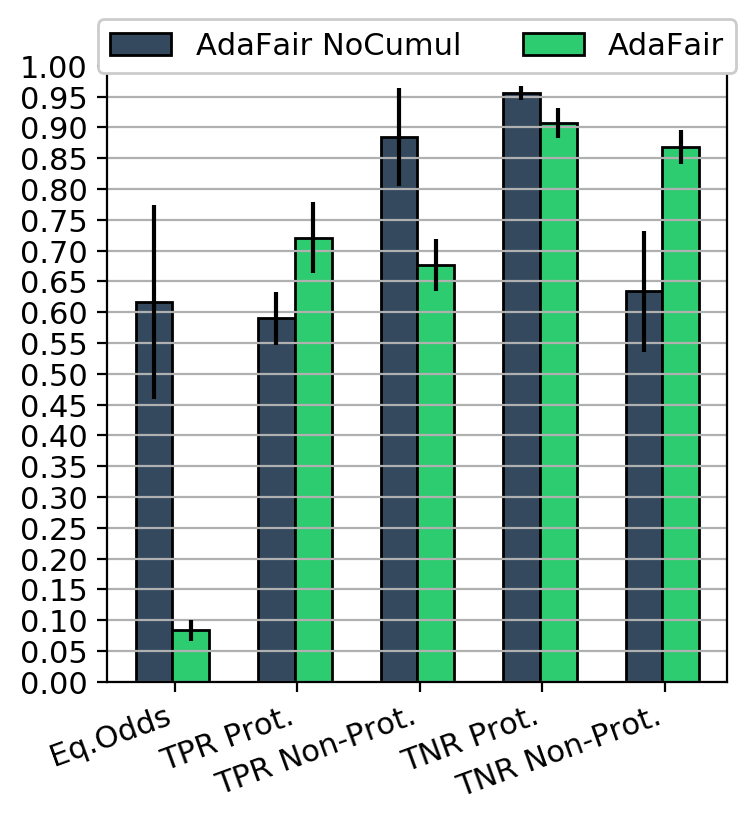}
 \caption{Adult census}
 \end{subfigure}
 \centering
 \begin{subfigure}[t]{0.24\textwidth}
 \centering
 \includegraphics[width=1.04\columnwidth]{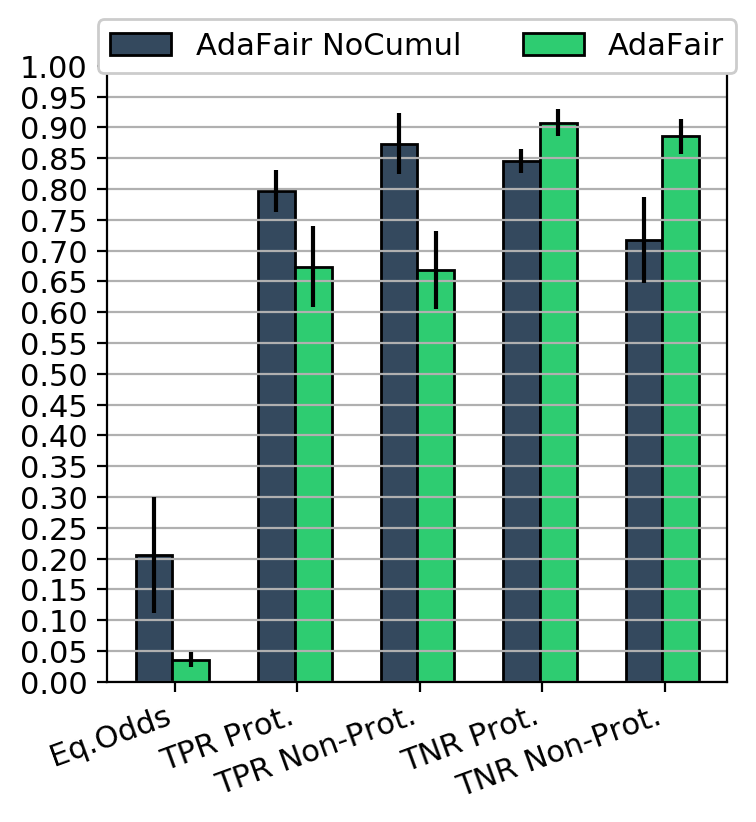}
 \caption{Bank}
 \end{subfigure}
 \begin{subfigure}[t]{0.24\textwidth}
 \centering
 \includegraphics[width=1.04\columnwidth]{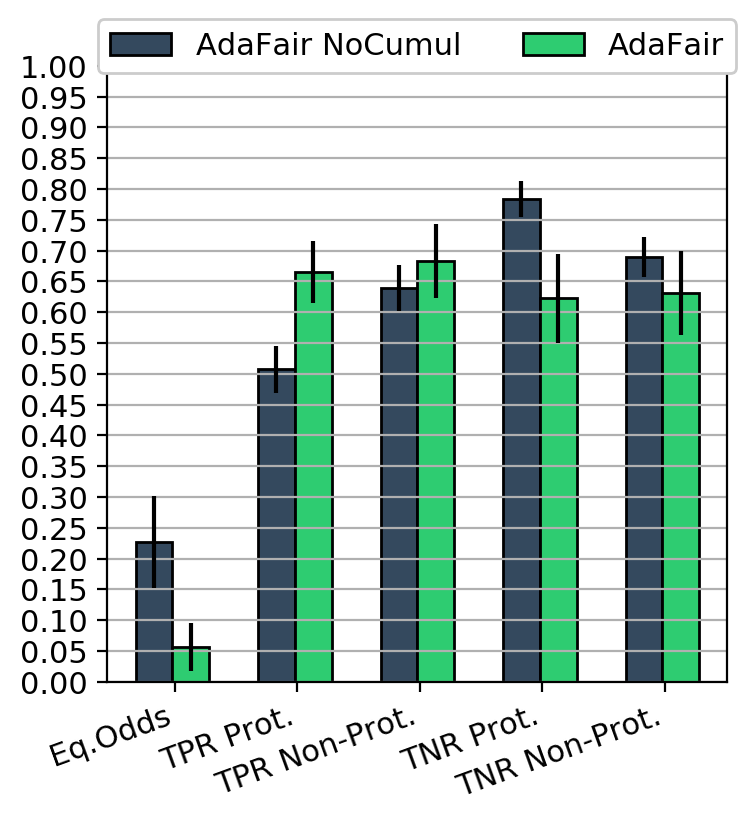}
 \caption{Compass} 
 \end{subfigure}
 \begin{subfigure}[t]{0.24\textwidth}
 \centering
 \includegraphics[width=1.04\columnwidth]{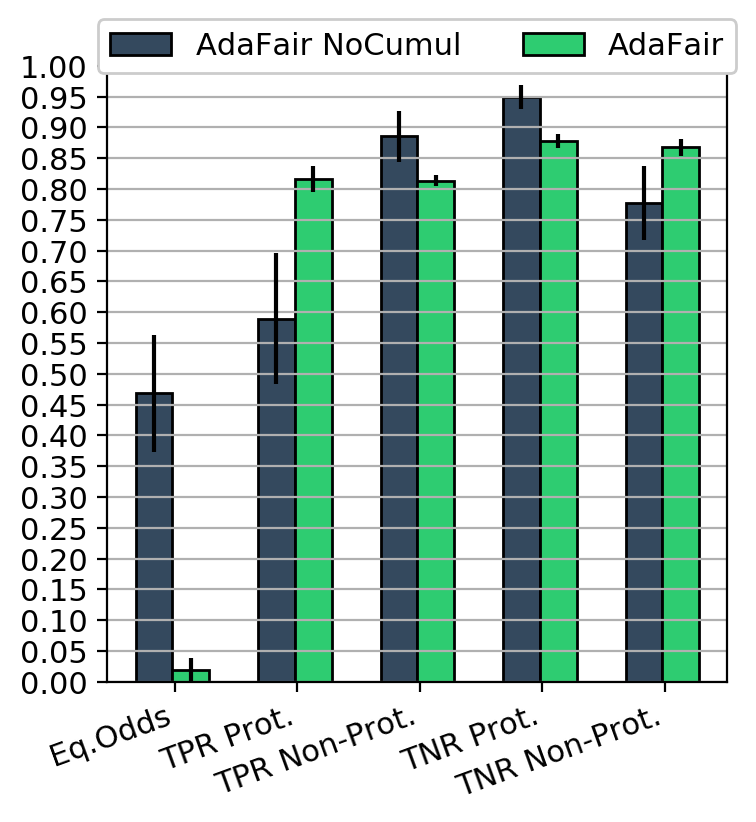}
 \caption{KDD census} 
 \end{subfigure}
 \caption{
Effect of cumulative fairness: 
\ourmethod vs \ourmethod NoCumul} 
 \label{fig:single_performance}
\end{figure*}

Their predictive and fairness performance for the different datasets is shown in Figure~\ref{fig:single_performance}. Overall, AdaFair NoCumul method results in poor fairness performance with very high $Eq.Odds$ values compared to \ourmethod.
In particular, we observe an increase of 52\%$\uparrow$ for the adult dataset, an increase of 15\%$\uparrow$ for bank and compass datasets and an increase of 45\%$\uparrow$ for the kdd census dataset. A closer look at the individual TPR, TNR scores shows that regarding TPR, the scores of the protected group are lower whereas w.r.t. TNR, the scores of the protected group are higher. That is, more protected instances are rejected (low TPR, high TNR). Moreover, standard deviation for the non-cumulative version is higher than \ourmethod, indicating AdaFair NoCumul is not stable.
It appears that a cumulative notion of fairness based on the current ensemble composition is better than a non-cumulative approach that considers solely the fairness behavior of the current/last weak learner.

\begin{figure*}[htp!]
 \centering
 \begin{subfigure}[t]{0.24\textwidth}
 \centering
 \includegraphics[width=1.04\columnwidth]{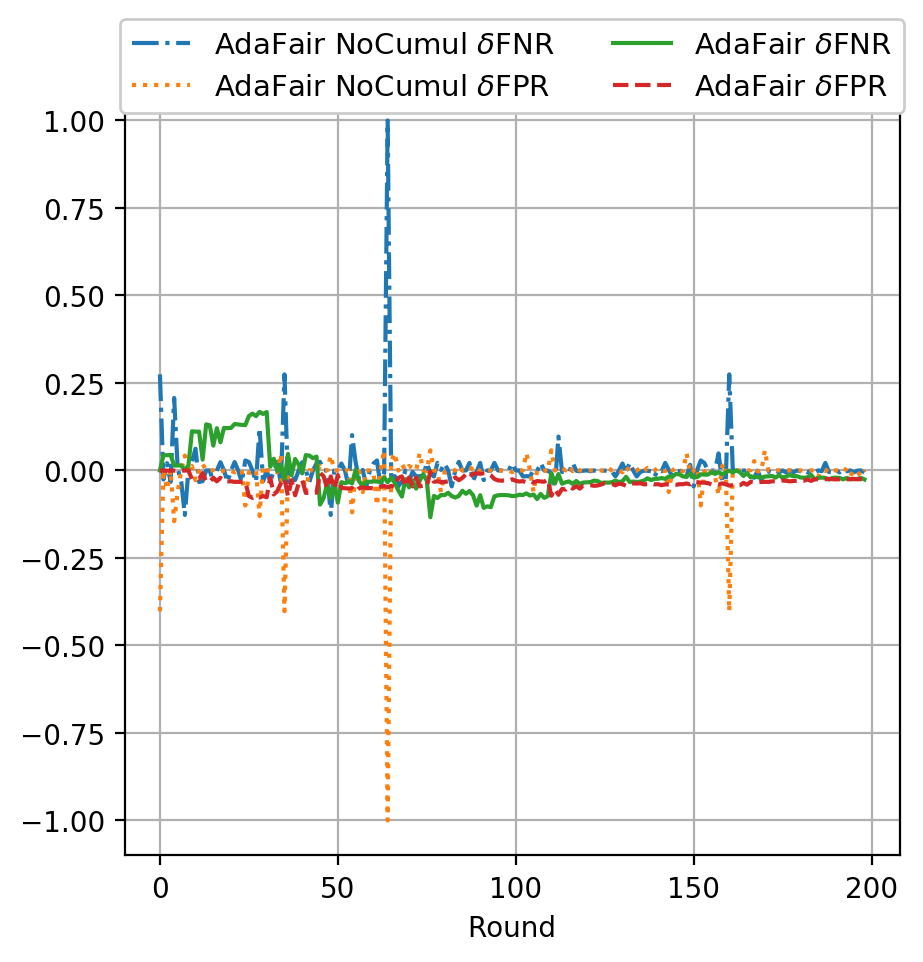}
 \caption{Adult census}
 \end{subfigure}
 \centering
 \begin{subfigure}[t]{0.24\textwidth}
 \centering
 \includegraphics[width=1.04\columnwidth]{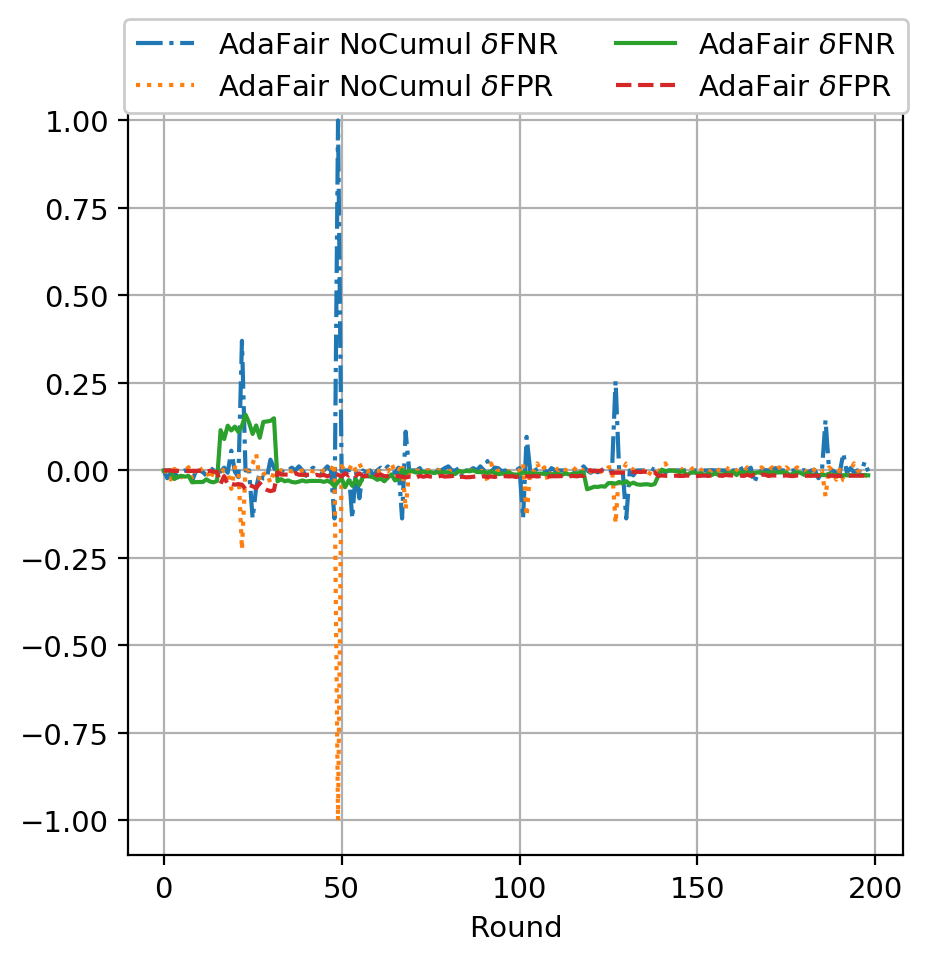}
 \caption{Bank}
 \end{subfigure}
 \begin{subfigure}[t]{0.24\textwidth}
 \centering
 \includegraphics[width=1.04\columnwidth]{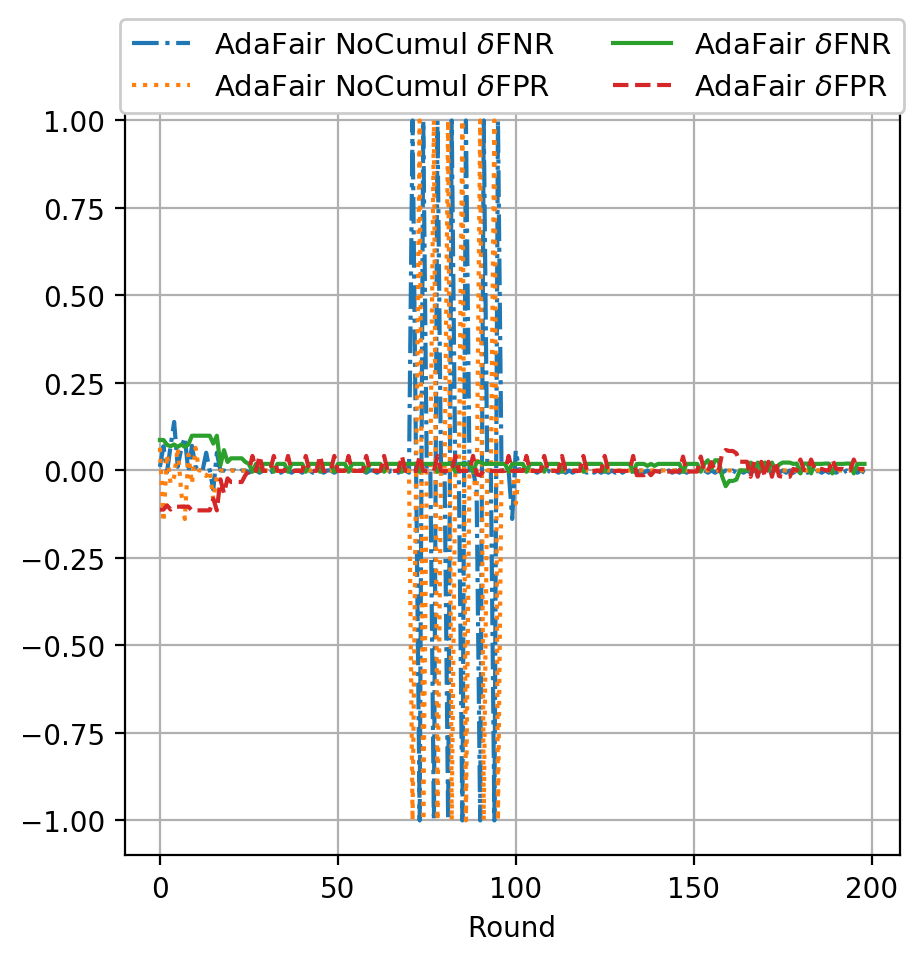}
 \caption{Compass} 
 \end{subfigure}
 \begin{subfigure}[t]{0.24\textwidth}
 \centering
 \includegraphics[width=1.04\columnwidth]{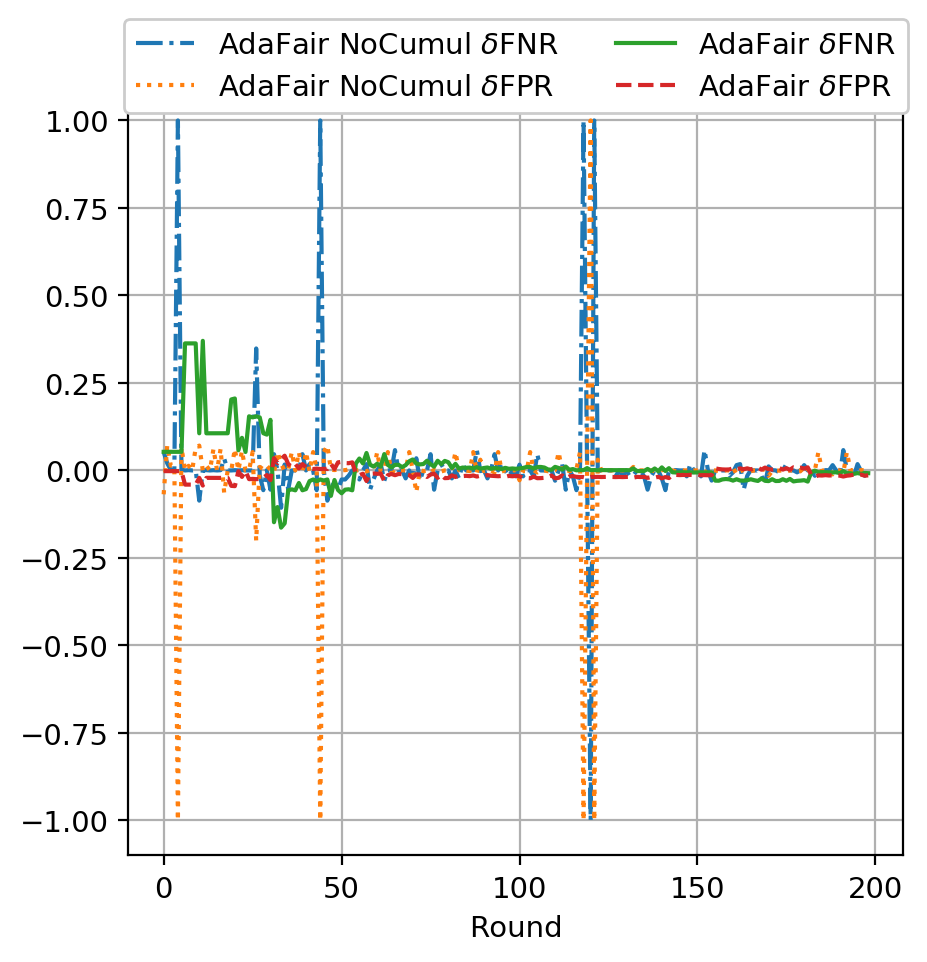}
 \caption{KDD census} 
 \end{subfigure}
 \caption{Costs per boosting round: \ourmethod vs \ourmethod NoCumul} 
 \label{fig:round_costs}
\end{figure*}

Except for the overall ensemble behavior, we also show the behavior per round. In particular, in Figure~\ref{fig:round_costs} we compare the per round $\delta FNRs$ and $\delta FPRs$ of the two approaches. Recall that $\delta FNR$ and $\delta FPR$ define the extra weighting/cost $u$ related to fairness that affects the weighting of the instances for the next round (see Section~\ref{sec:AdaFair_reweighting}).
Costs of AdaFair NoCumul exhibit a high fluctuation. On the contrary, the costs for our \ourmethod are smoother and converge after a sufficient number of rounds to a certain range $[-0.05, 0.05]$. That means that our method mitigates discrimination over the early rounds. This results further confirms that the cumulative definition of fairness is superior to a non-cumulative approach.


\subsection{The effect of confidence scores}
\label{sec:margins}
\begin{figure*}[h]
  \begin{subfigure}[t]{\textwidth}
  \centering
 \includegraphics[width=.73\columnwidth]{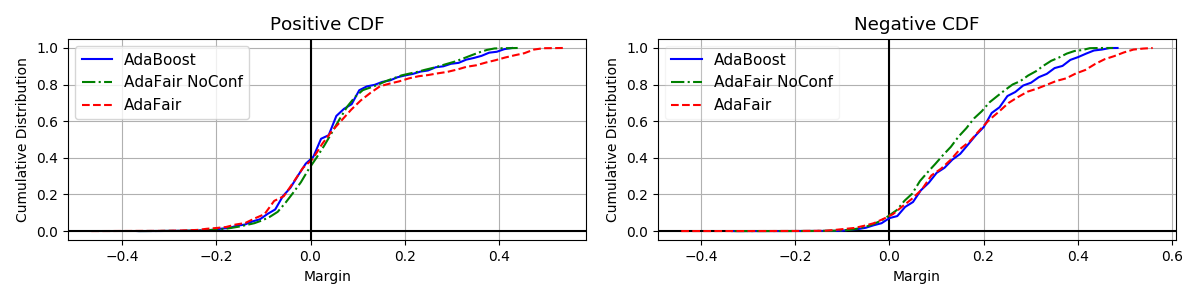}\vspace{-0.3cm}
 \caption{Adult census} 
 \end{subfigure}
  \begin{subfigure}[t]{\textwidth}
  \centering
 \includegraphics[width=.73\columnwidth]{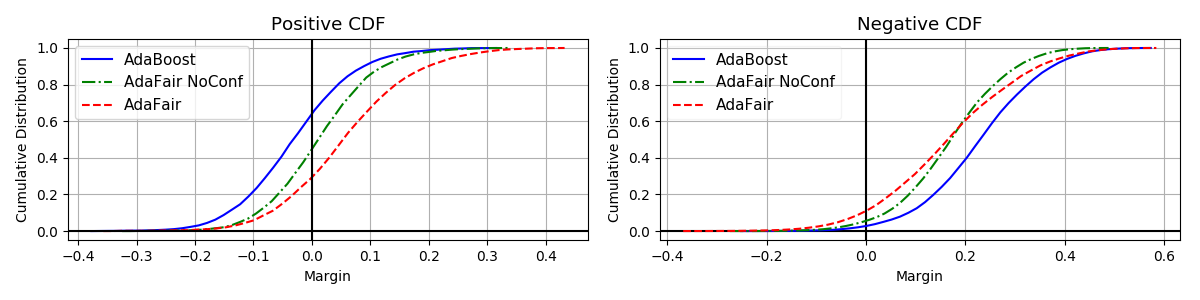}\vspace{-0.3cm}
 \caption{Bank} 
 \end{subfigure}
 \centering
 \begin{subfigure}[t]{\textwidth}
  \centering
 \includegraphics[width=.73\columnwidth]{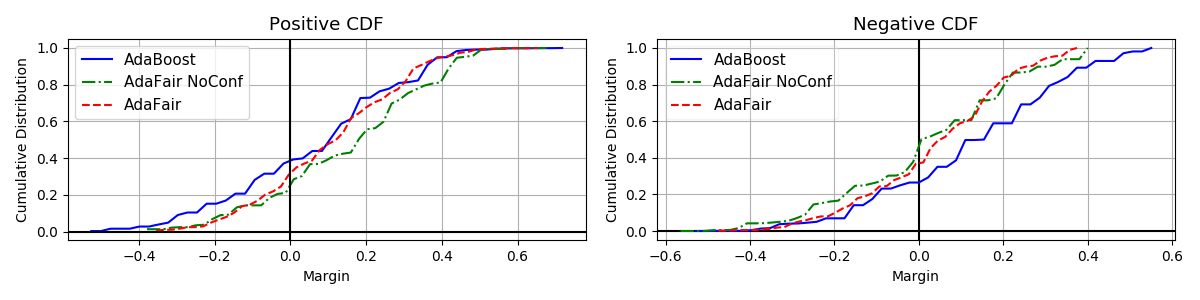}\vspace{-0.3cm}
 \caption{Compass}
 \end{subfigure}
  \begin{subfigure}[t]{\textwidth}
  \centering
 \includegraphics[width=.73\columnwidth]{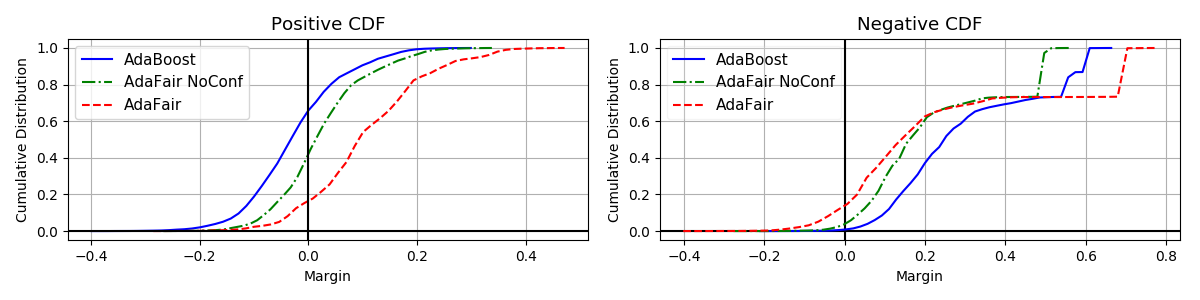}\vspace{-0.3cm}
 \caption{KDD census} 
 \end{subfigure}
 \caption{Effect of confidence scores per dataset}
 \label{fig:cdf_and_weights}
\end{figure*}

Our \ourmethod utilizes confidence scores in the instance weight estimation (c.f., Section~\ref{sec:AdaFair_reweighting}) in contrast to vanilla AdaBoost. 
To demonstrate the impact of confidence scores, we compare \ourmethod to a variation that does not use confidence scores (Algorithm~\ref{alg:method} line 2g, confidence score $\hat{h}$ is removed); we refer to this as \ourmethod NoConf, hereafter. We also include vanilla AdaBoost in the comparison. We evaluate the three approaches w.r.t. their cumulative margins. Comparing to accuracy and balanced accuracy that rely only on correctly and incorrectly classifications, the margins also reveal information about how confident are the predictions of the different methods~\cite{schapire2013explaining}. The cumulative distribution of margins for each dataset and per class is provided in Figure~\ref{fig:cdf_and_weights}. The margin values lie in the $[-1, 1]$ range; values close to $-1$ indicate misclassifications with high confidence whereas values close to $1$ indicate correct classifications with high confidence. The bulk of the distribution is located where the curve increases abruptly.

Based on the results, we observe that for the positive class our \ourmethod is more confident comparing to the other approaches (its curve is shifted to the right) and this holds for all datasets except for compass where there is no clear winner between \ourmethod and \ourmethod NoConf - recall that this dataset is balanced. Also, we can observe that \ourmethod has the lowest errors in the positive class and the misclassifications in this class have lower confidence values comparing to the other methods. 
Regarding the negative class, we can observe that AdaBoost achieves the best performance compared to \ourmethod and \ourmethod NoConf, since it focuses solely on overall accuracy, thus it learns better the majority class.

The differences in the performances of the different methods increase with the dataset imbalance; for example for the \emph{Bank dataset}, \ourmethod has significantly less misclassifications in the positive class comparing to \ourmethod NoConf ($15\% \downarrow$) and AdaBoost ($36\% \downarrow$), and only small increase in the misclassifications of the negative class, $5\% \downarrow$ and $7\% \downarrow$, respectively.
Similarly, for the \emph{KDD dataset}, our \ourmethod has $23\% \downarrow$ and $46\% \downarrow$ less misclassifications than \ourmethod NoConf and AdaBoost, respectively, while the increase in error at the negative class is $12\% \downarrow$ and $14\% \downarrow$, respectively. 
In this case, AdaBoost classifies correctly almost all the negative (majority) instances, but its performance on the minority class is the worst (it has the leftmost curve).

To conclude, considering confidence scores for instance weight estimation, the ensemble's margins increase especially in case of class imbalance. 
Instead of ``boosting'' all misclassified instances equally, this approach differentiates based on the confidence score of the misclassification, s.t. misclassified instances close to the decision boundary are weighted less than those that are further away.

\subsection{The effect of balanced error}
\label{sec:exp_parameter_c}
\begin{figure*}[htp]
  \centering
  \begin{subfigure}[t]{0.24\textwidth}
    \centering
  \includegraphics[width=1.0025\columnwidth]{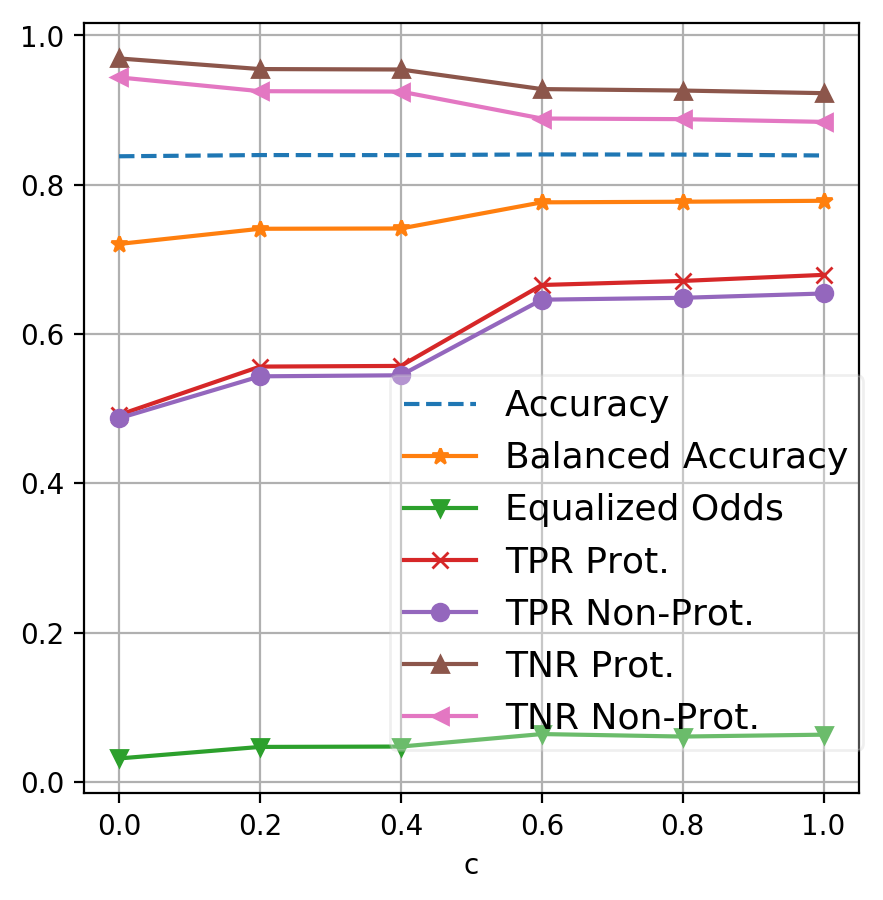}
    \caption{Adult census}
  \end{subfigure}
  \begin{subfigure}[t]{0.24\textwidth}
    \centering
  \includegraphics[width=1.0025\columnwidth]{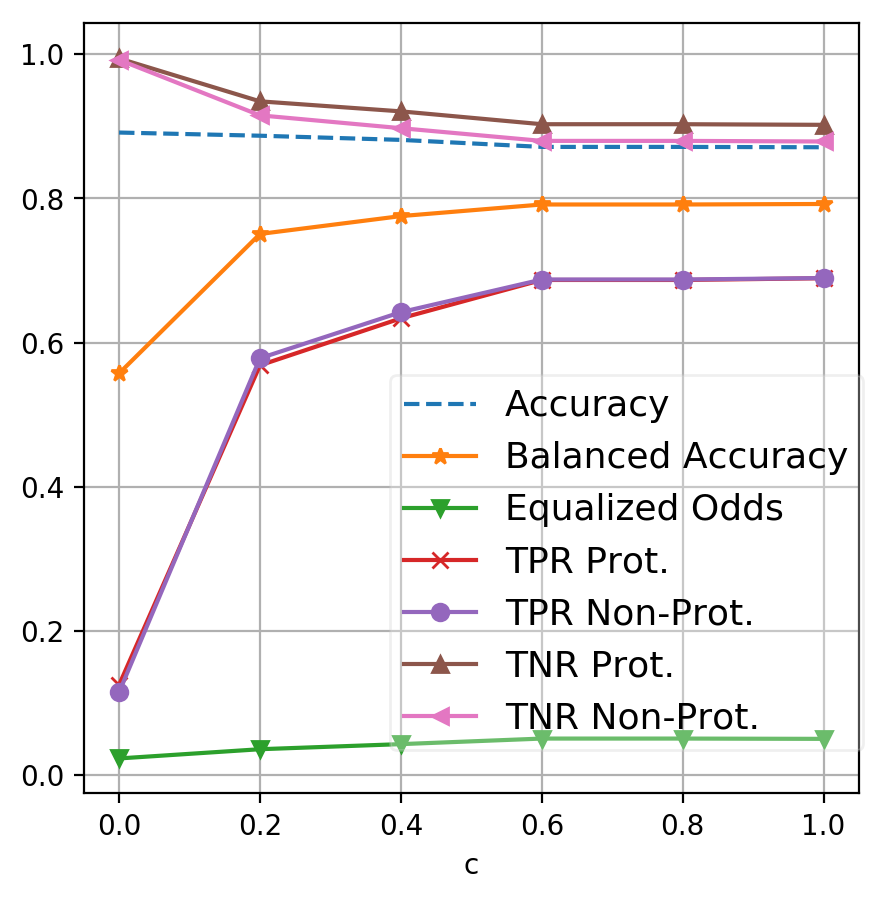}
    \caption{Bank}
  \end{subfigure}
  \begin{subfigure}[t]{0.24\textwidth}
    \centering
    \includegraphics[width=1.0025\columnwidth]{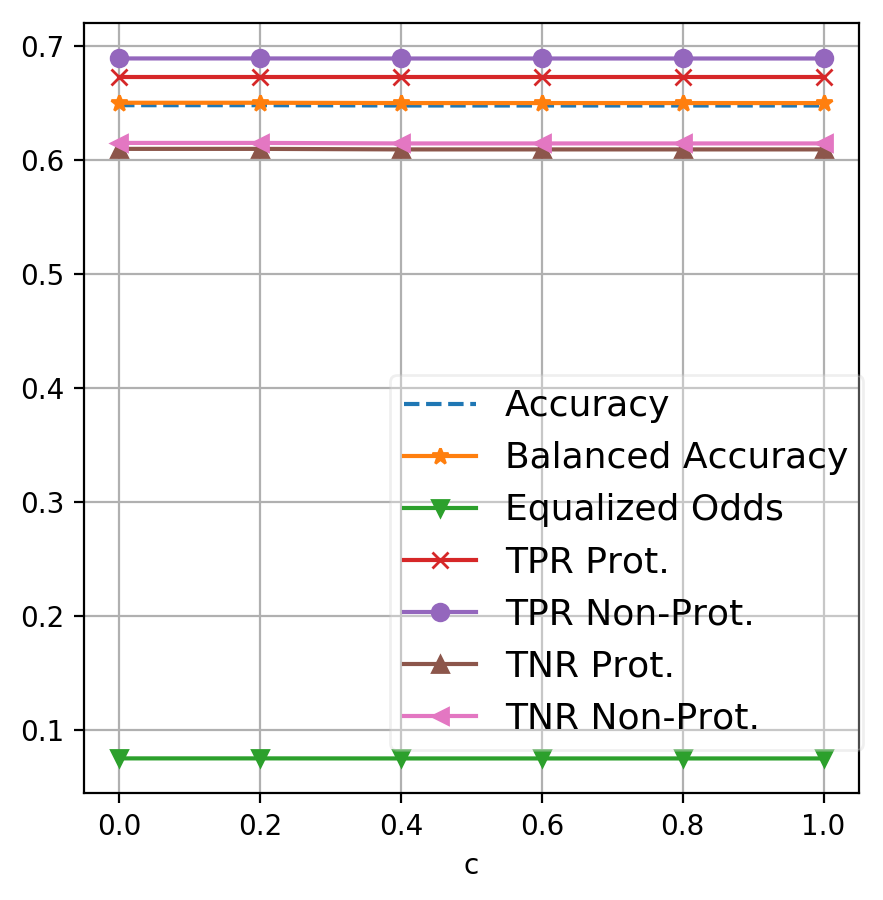}
    \caption{Compass}
  \end{subfigure}
    \begin{subfigure}[t]{0.24\textwidth}
    \centering
    \includegraphics[width=1.0025\columnwidth]{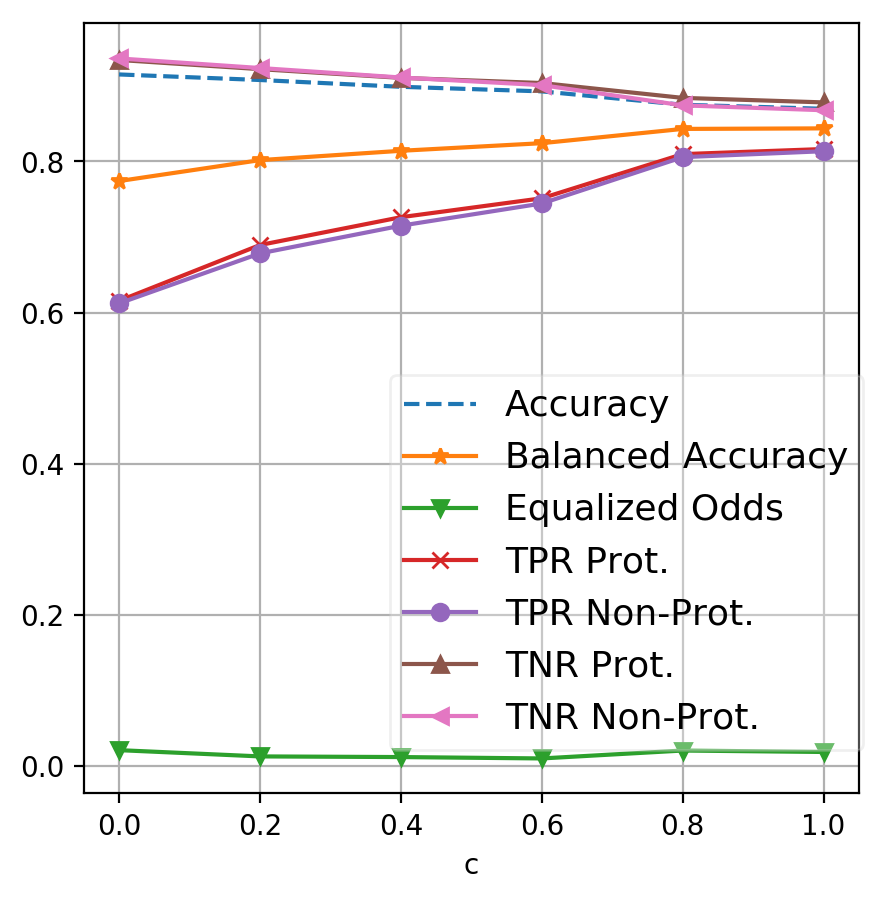}
    \caption{KDD census}
  \end{subfigure}
  \caption{The effect of balanced error (here we report on accuracy and balanced accuracy instead)}
  \label{fig:impact_of_c}
\end{figure*}

Our \ourmethod is able to achieve parity even when it does not optimize for balanced error rate because of the cumulative fairness and confidence scores that alter the data distribution during training in the direction of fairness. In this section, we show that varying parameter $c$, which alternates the objective goal (Equation~\ref{eq:argmin}), has a strong positive effect on TPRs while only slightly affecting TNRs. 

In particular, in Figure~\ref{fig:impact_of_c}, we plot accuracy and fairness related measures for different values of $c \in [0, 1]$. For $c=1$, the balanced error is optimized (our proposed \ourmethod). For $c=0$, the error rate is optimized. Values in-between (we use a step of 0.2 for $c$) correspond to different combinations of balanced error and error rate. Values are averaged over 10 random splits.
Note that different values of $c$ result into different \ourmethod sizes, i.e., the sub-sequence of weak learners in the ensemble varies with $c$. 

As we can see, for all imbalanced datasets (adult census, bank, kdd census) the balanced accuracy increases with $c$. For $c=0$ (only error rate is considered) the TPRs for both groups are very low. The TPRs increase with $c$, reaching their best values at $c=1$, i.e., when balanced accuracy is considered.
TNRs decrease with $c$, though their decrease is lower than the increase of TPRs. This again supports our previous findings that \ourmethod achieves parity between the two groups for both TPRs and TNRs while achieving high TPRs, on the contrary to Zafar et al. and Krasanakis et al. (c.f., Section~\ref{sec:evaluation}). For Compass, accuracy and balanced accuracy are very close and therefore no significant differences are observed by varying $c$.

More precisely, a comparison of the results for $c=0$ (error rate) and $c=1$ (balanced error rate) shows i) for the adult dataset, 18\%$\uparrow$ and 15\%$\uparrow$ increase in TPRs (for $s$, $\bar{s}$ groups respectively) and only 5\%$\downarrow$ reduction in TNRs (for both groups); 
ii) for the bank dataset, 56\%$\uparrow$ increase in TPRs (for both groups) and only 8\%$\downarrow$ and 9\%$\downarrow$ decrease in TNRs (for $s$ and $\bar{s}$,respectively)
and iii) for the kdd dataset,  20\%$\uparrow$ growth in TPR (for both groups) and only  5\%$\downarrow$ and 6\%$\downarrow$ reduction in TNRs (for $s$ and $\bar{s}$, respectively).

To conclude, directly tackling the imbalance via the balanced error is more effective for fairness-aware classification.

\section{Conclusions and Future Work}
\label{sec:conclusions}
In this work, we have propose \ourmethod, a fairness-aware boosting approach that adapts AdaBoost to fairness by changing the data distribution at each round based on the notion of cumulative fairness that evaluates per round the fairness-related behavior of the thus far built ensemble and adjusts accordingly the weight/cost of the per round discriminated group.

Our experiments, on four real-world datasets, show a substantial difference in TPRs compared to current fairness aware state-of-the-art approaches. We have seen, in cases of severe and extreme class-imbalance, that AdaFair can achieve parity among protected and non-protected group and at the same time maintain significantly better classification performance compared to other approaches. Moreover, in cases of extreme class-imbalance, AdaFair is able to outperform methods which focus solely on class-imbalance. In cases of class-balanced datasets, such as compass, AdaFair achieves parity similar to other fairness aware methods. 


Moreover, AdaFair is able to achieve parity even when it does not optimize for balanced error, however this impacts its performance in terms of balanced error rate i.e., low TPRs, high TNRs and low Eq.Odds. In addition, from our comparisons of cumulative versus non-cumulative fairness, we come to the conclusion that cumulative fairness is without doubt superior to non-cumulative fairness. Furthermore, adapting confidence scores in the weight estimation process makes AdaFair more confident compared to vanilla AdaBoost.

A possible extension of our method is the online selection of the optimal number of rounds $\theta$ for the ensemble, to replace our post-training search for the optimal $\theta$ parameter. Moreover, we plan to work on the theoretical aspects of our method, in particular regarding its convergence properties.


\begin{acks}
The work was funded by the German Research Foundation (DFG) project OSCAR (Opinion Stream Classification with Ensembles and Active leaRners) and inspired by the Volkswagen Foundation project BIAS ("Bias and Discrimination in Big Data and Algorithmic Processing. Philosophical Assessments, Legal Dimensions, and Technical Solutions") within the initiative "AI and the Society of the Future"; the last author is a Project Investigator for both of them.
\end{acks}

\bibliographystyle{ACM-Reference-Format}
\bibliography{bibliography}

\end{document}